\documentclass[letterpaper]{article} 
\usepackage{aaai2026}  
\usepackage{times}  
\usepackage{helvet}  
\usepackage{courier}  
\usepackage[hyphens]{url}  
\usepackage{graphicx} 
\usepackage{natbib}  
\usepackage{caption} 
\usepackage{algorithm}
\usepackage{algorithmic}

\expandafter\let\csname equation*\endcsname\relax
\expandafter\let\csname endequation*\endcsname\relax
\usepackage{amssymb}
\usepackage{amsmath}

\usepackage{booktabs}
\usepackage{multirow}
\usepackage{siunitx}  
\usepackage{booktabs}
\usepackage{multirow}
\usepackage{siunitx}  
\usepackage{subcaption}

\newcommand{\ie}{\textit{i.e.}}
\newcommand{\eg}{\textit{e.g.}}

\usepackage{newfloat}
\usepackage{listings}
\DeclareCaptionStyle{ruled}{labelfont=normalfont,labelsep=colon,strut=off} 
\floatstyle{ruled}
\newfloat{listing}{tb}{lst}{}
\floatname{listing}{Listing}
\title{WaveTuner: Comprehensive Wavelet Subband Tuning for Time Series Forecasting}
\author{
    Yubo Wang\textsuperscript{\rm 1},
    Hui He\textsuperscript{\rm 1},
    Chaoxi Niu\textsuperscript{\rm 2}
    Zhendong Niu\textsuperscript{\rm 1}
}
\affiliations{
    \textsuperscript{\rm 1} Beijing Institute of Technology\\
    \textsuperscript{\rm 2} University of Technology Sydney\\

}

\usepackage{bibentry}

\begin{document}

\maketitle

\begin{abstract}

Due to the inherent complexity, temporal patterns in real-world time series often evolve across multiple intertwined scales, including long-term periodicity, short-term fluctuations, and abrupt regime shifts. While existing literature has designed many sophisticated decomposition approaches based on the time or frequency domain to partition trend–seasonality components and high–low frequency components, an alternative line of approaches based on the wavelet domain has been proposed to provide a unified multi-resolution representation with precise time–frequency localization. However, most wavelet-based methods suffer from a \textit{persistent bias} toward recursively decomposing only low-frequency components, severely underutilizing \textit{subtle yet informative} high-frequency components that are pivotal for precise time series forecasting. To address this problem, we propose WaveTuner, a \textit{Wave}let decomposition framework empowered by full-spectrum subband \textit{Tun}ing for time series forecasting. Concretely, WaveTuner comprises two key modules: (i) \textit{Adaptive Wavelet Refinement module}, that transforms time series into time-frequency coefficients, utilizes an adaptive router to dynamically assign subband weights, and generates subband-specific embeddings to support refinement; and (ii) \textit{Multi-Branch Specialization module}, that employs multiple functional branches, each instantiated as a flexible Kolmogorov–Arnold Network (KAN) with a distinct functional order to model a specific spectral subband.
Equipped with these modules, WaveTuner comprehensively tunes global trends and local variations within a unified time–frequency framework. Extensive experiments on eight real-world datasets demonstrate WaveTuner achieves state-of-the-art forecasting performance in time series forecasting.

\end{abstract}

\section{Introduction}



Time series forecasting, which aims to infer future values from temporal patterns of historical observations, plays a pivotal role in a wide range of real-world applications, such as transportation management~\cite{cirstea2022towards}, inventory optimization~\cite{seyedan2023order}, and climate modeling~\cite{haq2022cdlstm}. In recent years, various deep learning models based on diverse architectures—such as RNNs~\cite{amalou2022multivariate}, CNNs~\cite{mehtab2022analysis}, Transformers~\cite{woo2024unified}, and MLPs~\cite{zeng2023transformers}—have gained significant attention and driven notable progress for time series forecasting~\cite{lim2021time}.

\begin{figure}[!t]
\centering
\subfloat[Entropy–based two-level wavelet packet decomposition]
{\includegraphics[width=1.0\columnwidth]{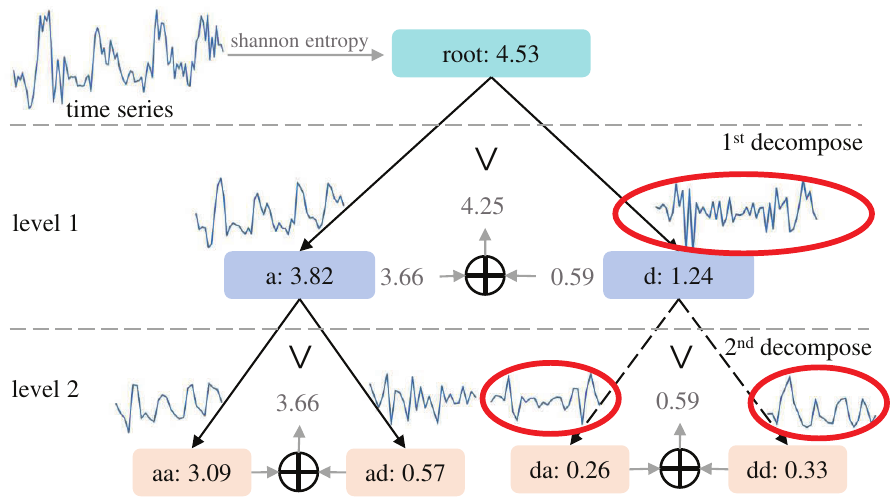}}\\

\subfloat[Wavelet Spectra Visualization of $a$ (\textbf{Left}) and $d$ (\textbf{Right})]
{\includegraphics[width=1.0\columnwidth]{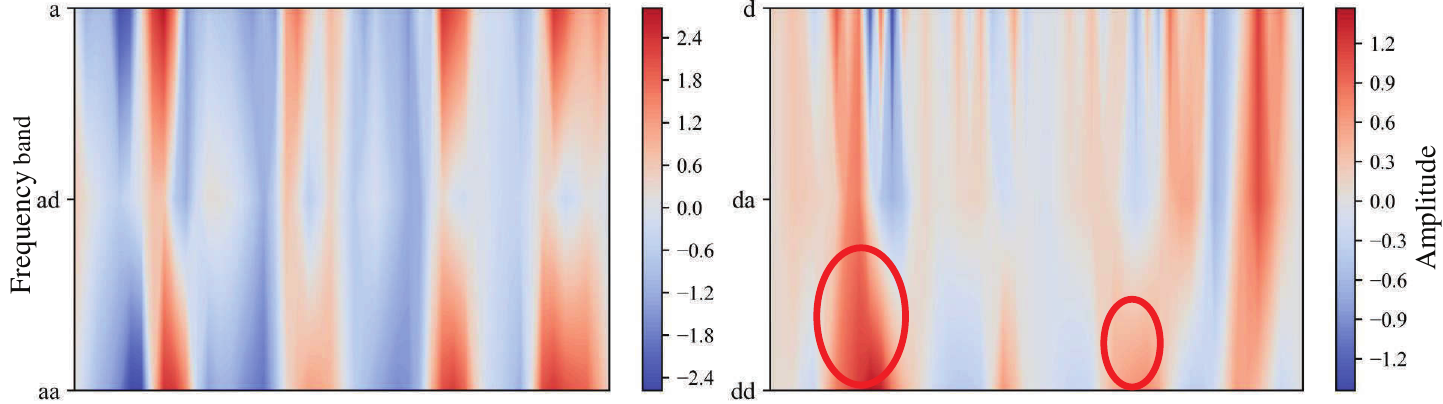}}
\\

\caption{The critical role of analyzing high-frequency components in wavelet decomposition.
}

\label{fig1}
\end{figure}

Despite these impressive advances, modeling time series remains fundamentally challenging due to the intrinsic complex nature of the real world, where temporal patterns unfold across multiple entangled scales, including long-term periodicity, short-term fluctuations, abrupt regime shifts, etc~\cite{DBLP:conf/aaai/ZhangPADATHVM25,DBLP:conf/kdd/Piao0MMS24}. 
To tackle such complex temporal patterns, a compelling strategy is to leverage prior knowledge to decompose time series into trend and seasonal components~\cite{wu2021autoformer,DBLP:conf/icml/ZhouMWW0022,zeng2023transformers}, further enriched with multi-scale refinements that capture cross-scale interactions~\cite{DBLP:conf/iclr/WangWSHLMZ024}, or into chunks with different period lengths~\cite{wu2022timesnet}.
Concurrently, the frequency domain has emerged as a powerful alternative to conventional time-domain approaches by providing global view and energy compaction, two advantaged properties inaccessible in the time domain~\cite{DBLP:conf/nips/YiZFWWHALCN23}, prompting a surge of interest in decomposing time series into high- and low-frequency components~\cite{piao2024fredformer,huang2025timekan,DBLP:conf/iclr/WuQL0HGXY25}. Nevertheless, frequency-based decomposition remains fundamentally constrained in capturing time-sensitive patterns that evolve dynamically over time. Surpassing the inherent limitations of pure time- or frequency-domain approaches, the wavelet domain is rapidly gaining momentum for its unique ability to unify time and frequency analysis, yielding multi-resolution and time-sensitive representations with strong localization across both domains\cite{guo2022review}.

However, wavelet-empowered forecasters are still suffering from a persistent \textit{bias} toward recursively decomposing only low-frequency signals (\ie, approximation coefficients), rendering them particularly \textit{vulnerable} to high-frequency signals (\ie, detail coefficients)---\textit{subtle yet informative} components for accurately forecasting time series.
Such bias severely undermines the full potential of the wavelet domain. To highlight the importance of high-frequency signals, Figure \ref{fig1} (a) illustrates a two-level optimal subband tree guided by Shannon entropy. 
Although the high-frequency band $d$ is typically underexplored by existing methods~\cite{yi2024filternet}, it exhibits a high entropy of 1.24, suggesting the presence of rich structural information. Upon further decomposition of $d$, the component of $dd$ still exhibits pronounced periodic patterns (see red circles), indicating that $d$ retains entangled yet structured temporal patterns that merit deeper decomposition for more effective modeling. Additionally, the wavelet spectra of $da$ and $dd$ exhibit strong time-localized information (see Figure \ref{fig1} (b)), on par with those observed in the $aa \leftarrow a \rightarrow ad$ branch. This observation further reinforces the necessity of deeper decomposition to isolate more informative time-frequency characteristics.

To address the aforementioned issues, we propose WaveTuner, a \textit{Wave}let decomposition framework empowered by full-spectrum subband \textit{Tun}ing for effective time series forecasting. The core idea of WaveTuner is to adaptively focus on high-frequency detail coefficients across multi-resolution wavelet subbands, facilitating the discovery of optimal subband routing patterns tailored to each time series input. 
Specifically, we introduce the Adaptive Wavelet Refinement (AWR) module, which transforms time series into time-frequency coefficients and utilizes an adaptive router to dynamically assign subband weights, enabling subband-specific refinement that enhances the model’s ability to capture localized frequency dynamics. These coefficients are further refined to model inter-variable dependencies via hardware-friendly linear layers with residual connections, yielding finer time-frequency representations that empower WaveTuner to capture more informative and discriminative patterns across diverse spectral bands. 
Inspired by the exceptional data-fitting capacity of Kolmogorov-Arnold Networks (KAN), we design the Multi-branch Specialization (MBS) module, where each branch---instantiated as a KAN of different functional order---is specialized for a specific spectral band. This design aligns model complexity with frequency characteristics: low-frequency subbands benefit from smoother, low-order functions to capture global trends, while high-frequency subbands require higher-order expressiveness to model rapidly changing local patterns.

Our contributions can be summarized as follows:
\begin{itemize}
    \item We reveal a strong bias toward high-frequency components of current wavelet-based solutions, and propose WaveTuner, a novel wavelet decomposition framework that enables full-spectrum subband tuning for effective time series forecasting.
    \item We introduce the AWR module to transform time series into time-frequency coefficients and dynamically assign importance weights to each subband, enabling adaptive emphasis across the frequency spectrum.
    \item We then introduce the MBS module to leverage frequency-specific KAN-based subnetworks with varying functional orders to align model expressiveness with spectral characteristics.
    \item Extensive experiments on eight forecasting benchmarks demonstrate the superiority of our model over state-of-the-art methods.
\end{itemize}

\section{Related Works}


\paragraph{Deep Time Series Forecasting.}
Recent advances in TSF span various architectural paradigms, including CNN-, RNN-, Transformer-, and MLP-based methods. Early models such as DeepAR~\cite{salinas2020deepar} and SCINet~\cite{liu2022scinet} utilized RNN and CNN structures to capture temporal dependencies, but struggled with long-range forecasting. Transformer-based models, such as Informer~\cite{zhou2021informer}, Autoformer~\cite{wu2021autoformer}, Crossformer~\cite{zhang2023crossformer}, and iTransformer~\cite{liu2023itransformer}, have significantly improved long-horizon prediction through sparse attention, series decomposition, and patch-based representations~\cite{nie2022time}. More recently, MLP-based approaches~\cite{tang2025unlocking} have gained attention due to their architectural simplicity and competitive performance. DLinear~\cite{zeng2023transformers}, RLinear~\cite{li2023revisiting}, and TimeMixer~\cite{wang2024timemixer} employ trend-remainder decomposition, MLP or multi-scale mixing strategies. 
PatchMLP~\cite{tang2025unlocking} and TVNet~\cite{li2025tvnet} further show that patching enhances local temporal pattern modeling. 
Beyond time-domain modeling, frequency-aware methods have emerged as an effective alternative. FredFormer~\cite{piao2024fredformer}, FilterNet~\cite{yi2024filternet}, and ReFocus~\cite{yu2025refocus} leverage Fourier transforms to highlight mid- or high-frequency information. 
TimeKAN~\cite{huang2025timekan} combines frequency decomposition with Multi-order KANs~\cite{liu2024kan}, capturing nonlinear dynamics across different frequency bands. 
In contrast to conventional time- or frequency-domain approaches, WaveTuner extracts continuous wavelet subband features to construct multi-resolution representations that are both temporally sensitive and well-localized.

\paragraph{Wavelet Analysis in Time Series Modeling}
Unlike the Fourier transform, the wavelet transform offers time-frequency localization, enabling signal modeling at multiple resolutions. WPMixer~\cite{murad2025wpmixer} leverages wavelet-based multilevel decomposition and integrates patch mechanisms to model wavelet coefficients. AdaWaveNet~\cite{yu2024adawavenet} decomposes time series into seasonal and trend components and applies a lifting scheme to model the seasonal part. WaveletMixer~\cite{zhang2025waveletmixer} utilizes multi-resolution wavelet decomposition combined with a lightweight MLP-mixer architecture to enhance long-term multivariate time series forecasting. WFTNET~\cite{liu2024wftnet} combines Fourier transform and wavelet transform to explore global and local periods. WTFTP~\cite{zhang2023flight} combine the Wavelet Transformer with the encoder and decoder structures to predict aircraft trajectories. WaveRoRA~\cite{liang2024waverora} improves prediction performance by learning the relationship between frequency bands through the route attention mechanism. Wavelet transforms, when integrated with deep architectures such as CNNs and RNNs, have demonstrated substantial performance gains in various tasks~\cite{stefenon2023wavelet,stefenon2024hypertuned}. Moreover, wavelet packet transforms offer strong potential for time-series denoising, further enhancing robustness in downstream modeling~\cite{frusque2024robust}. 
In this paper, we explore a frequency-aware modeling approach that combines multi-resolution decomposition with an adaptive router to better capture the diverse dynamics in multivariate time series. 
 




\begin{figure*}[t]
\centering
\includegraphics[width=1.0\textwidth]{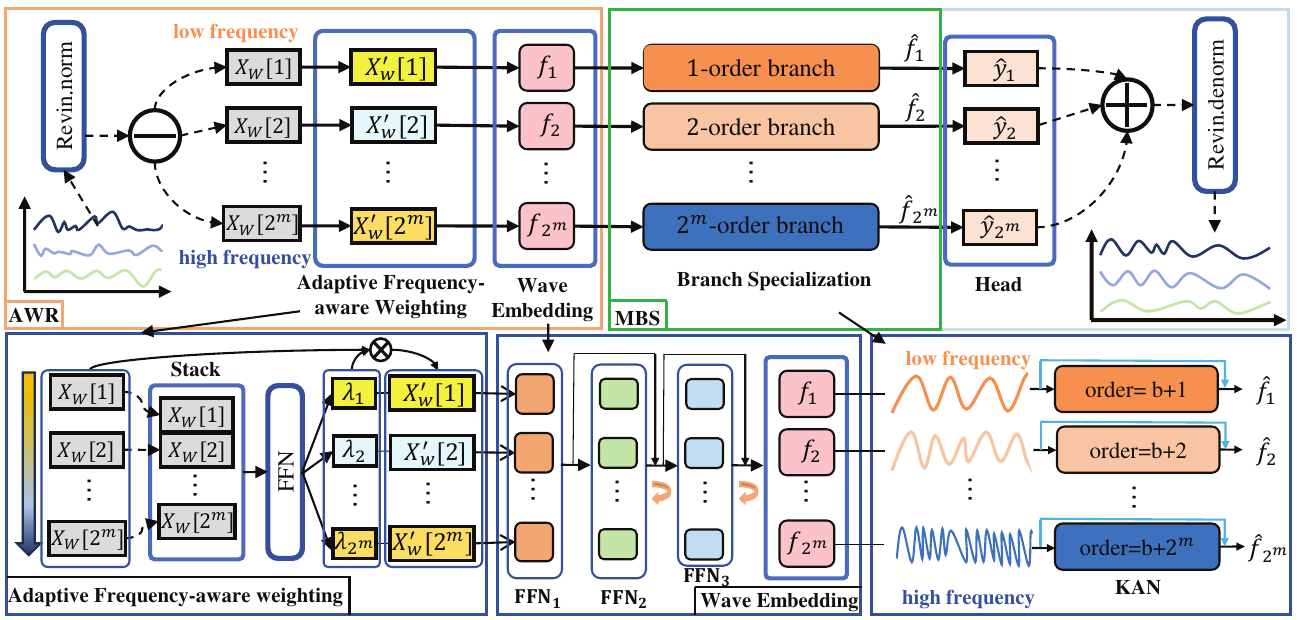} 
\caption{Framework of \textbf{WaveTuner}, composed of Adaptive Wavelet Refinement (AWR) and Multi-Branch Specialization (MBS). AWR applies wavelet packet decomposition, adaptive frequency-aware weighting, and wave embedding to highlight informative subbands. MBS assigns specialized branches to each band for frequency-specific modeling. Finally, the head module maps the results to the prediction horizon before reconstruction via inverse wavelet packet transform.}
\label{fig2}
\end{figure*}


\section{Methodology}
\subsection{Problem Formulation}
For time series forecasting, given an input multivariate time series \(X_{L}=\left \{ x_{t-L+1}, x_{t-L+2}, \dots,  x_{t-1}, x_{t}  \right \} \in \mathbb{R} ^{C\times L}\), the goal is 
to predict the future time series \(X_{T}=\left \{ x_{t+1}, x_{t+2}, \dots,  x_{t+T}  \right \} \in \mathbb{R} ^{C\times T}\), where \(x_{t} \in \mathbb{R} ^{1\times C}\) denotes a multivariate data point at time \(t\), \(C\) is the number of the variables, \(L\) and \(T\) are the length of look-back window and horizon window.

\subsection{Overview of WaveTuner}


The overall framework of WaveTuner is illustrated in Figure \ref{fig2}. WaveTuner consists of two core components: the Adaptive Wavelet Refinement (AWR) module and the Multi-Branch Specialization (MBS) module. The AWR module first applies wavelet packet decomposition to the normalized time series, generating multi-resolution wavelet coefficients. An adaptive routing mechanism is then employed to dynamically assign importance weights to different frequency subbands, enabling the model to selectively emphasize informative components across the spectrum. These weighted coefficients are further projected into a latent space to capture inter-variable dependencies at multiple frequency resolutions. Considering that low-frequency coefficients 
primarily encode long-term trends and high-frequency components reflect fine-grained fluctuations, the MBS module adopts a frequency-aware modeling strategy. It learns specialized representations for each subband to effectively capture the diverse and multi-scale temporal patterns inherent in multivariate time series.

\subsection{Adaptive Wavelet Refinement}
As shown in the upper-left part of Figure \ref{fig2}, AWR moduel consists Wavelet Packet Decomposition, Adaptive Frequency-aware Weighting and Wave Embedding.

\paragraph{Wavelet Packet Decomposition.} 
Since time series data often exhibit non-stationary behavior, RevIN~\cite{kim2021reversible} is employed for both normalization prior to wavelet packet decomposition (WPD) and denormalization after the reconstruction. 
WPD extends traditional wavelet decomposition by recursively applying both low-pass and high-pass filters at each level, enabling a full binary tree decomposition of the input. This approach preserves potentially valuable information contained in both low- and high-frequency components. Specifically, the normalized time series is transformed into a set of multi-level wavelet packet coefficients with enriched representational capacity, and the wavelet coefficients can be expressed as:
\begin{equation}
\begin{split}
\text{WPD}(X_{L}, \psi, m) = \left\{ X_w[i] \,\middle|\, i \in \{1, \dots, 2^m\} \right\} \\
= \left\{\text{band}_{j}^{(m)} \,\middle|\, \text{band}_{j} \in \{a,d\}^{m},\, j = 1,\dots,2^m \right\}\, ,
\end{split}
\end{equation}
where \(m\) denotes the number of decomposition levels, \(\psi \) represents the wavelet function, and \(n\) indicates the number of wavelet coefficients of \(X_{w}[i]\in \mathbb{R}^{C\times L_{i}} \) obtained after decomposition with \(L_{i}\) being the length of wavelet coefficent. Each subband is labeled as \(\text{band}_{j}^{(m)}\in\{a,d\}^{m}\), which represents a unique path from the root to a leaf node in the full binary tree.
For instance, with \(m=2\), the resulting subbands are \(aa\), \(ad\), \(da\), and \(dd\), corresponding to different combinations of approximation \(a\) and detail \(d\) operations applied across levels.

\paragraph{Adaptive Frequency-aware Weighting.} 
To effectively leverage the multi-band wavelet packet coefficients obtained through decomposition, we propose an adaptive frequency-aware weighting module, which functions as a soft routing mechanism. Instead of processing all frequency components equally, this module dynamically assigns importance scores to each subband based on the input characteristics and forecasting objective. In this way, it serves as an adaptive router that selectively emphasizes informative frequency components while suppressing irrelevant or noisy ones, guiding downstream modules to focus on task-relevant signals.

Specifically, for each frequency subband \(X_{w}[i]\) obtained from the wavelet packet decomposition, it is first processed through an average pooling operation to summarize its signal strength. This summary is then passed through a Feed-Forward Network (FFN) to output a weight \(\lambda_{i}\) for each band:
\begin{equation}
\lambda_{i} = \text{FFN}(\text{AvgPool}(X_{w}[i]))\, .
\end{equation}

The learned weight is then used to adjust the importance of each band as follows: 
\begin{equation}
X_{w}^{'}[i] = \lambda_{i}\cdot X_{w}[i]\, .
\end{equation}

By learning these weights in an end-to-end manner, the model effectively performs adaptive subband selection, where more relevant bands are emphasized while others are attenuated. This mechanism implicitly searches for an optimal subband tree structure without the need for hand-crafted frequency selection rules, improving both the flexibility and expressiveness of the model in the frequency domain. 

\paragraph{Wave Embedding.} 

To better model inter-variable relationships, the Wave Embedding module maps wavelet coefficients of the same frequency band across variables into high-dimensional space, where dependencies among variables can be effectively learned.
Specifically, the combined use of \(\text{FFN}_{1}^{i}\) and \(\text{FFN}_{2}^{i}\) with residual connections projects the coefficients to a rich feature representation:
\begin{equation}
X_{we}^{'}[i]=\text{FFN}_{1}^{i}(X_{w}^{'}[i])\, ,
\end{equation}
\begin{equation}
X_{we}^{'}[i]=\text{Norm}(\text{FFN}_{2}^{i}(X_{we}^{'}[i]) + X_{we}^{'}[i])\, .
\end{equation}

To capture the interactions across different frequency bands, \(\text{FFN}_{3}\) is applied on the transformed features:
\begin{equation}
f_{i}=\text{Re}(\text{Norm}(\text{Re}(\text{FFN}_{3}^{i}(X_{we}^{'}[i])+X_{we}^{'}[i])))\, ,
\end{equation}
where \(\text{FFN}_{i}\) denotes different feed-forward networks, and \(\text{Norm}(\cdot )\) represents layer normalization. The \(\text{Re}(\cdot)\) operation permutes the variable dimension to facilitate inter-variable modeling. The resulting \(f_{i}\) denotes the refined representation of the \(i\)-th frequency component, incorporating inter-variable dependencies. 

\subsection{Multi-Branch Specialization}
To model frequency-specific temporal dynamics, we further introduce the MBS module that assigns each wavelet subband a dedicated functional learner. Each branch employs a Chebyshev polynomial-based KAN with a chosen order. The polynomial order increases progressively with frequency, enabling low-frequency branches to capture smooth global trends, while high-frequency branches model fine-grained temporal variations. We adopt KAN as the functional learner for each subband due to its strong approximation and interpretability, achieved via learnable polynomial expansions. Compared with prior time-domain approaches~\cite{zeng2023transformers}, our frequency-domain specialization enhances both interpretability and modeling flexibility.

\begin{table*}[!t]
\centering
\setlength{\tabcolsep}{1mm}{
\begin{tabular}{cc|cc|cc|cc|cc|cc|cc|cc|cc}
\toprule

\multicolumn{2}{c|}{\textbf{Models}}
& \multicolumn{2}{c|}{\textbf{WaveTuner}} 
& \multicolumn{2}{c|}{WPMixer} 
& \multicolumn{2}{c|}{TimeKAN} 
& \multicolumn{2}{c|}{TimeMixer}
& \multicolumn{2}{c|}{FreTS}
& \multicolumn{2}{c|}{PatchTST} 
& \multicolumn{2}{c|}{TimesNet}
& \multicolumn{2}{c}{RLinear}

\\
\cmidrule(r){1-2}\cmidrule(r){3-4} \cmidrule(r){5-6} \cmidrule(r){7-8}\cmidrule(r){9-10} \cmidrule(r){11-12} \cmidrule(r){13-14}  \cmidrule(r){15-16} \cmidrule(r){17-18}
\multicolumn{2}{c|}{\textbf{Metrics}} 
& \textbf{MSE} & \textbf{MAE} 
& \textbf{MSE} & \textbf{MAE} 
& \textbf{MSE} & \textbf{MAE}  
& \textbf{MSE} & \textbf{MAE}
& \textbf{MSE} & \textbf{MAE}
& \textbf{MSE} & \textbf{MAE}
& \textbf{MSE} & \textbf{MAE}
& \textbf{MSE} & \textbf{MAE}
\\
\midrule

\multirow{4}{*}{\rotatebox{90}{ETTm1}}
& 96 & \underline{0.321} & \textbf{0.357}  & 0.332 & 0.362 &0.322 &0.361
&\textbf{0.320} &\underline{0.357} &0.335 &0.372 &0.329 &0.367 &0.338 &0.375  &0.355 &0.376
\\
& 192 & \underline{0.362} & \textbf{0.379} & 0.364 &\underline{0.379} & \textbf{0.357} &0.383
&0.367 &0.381  &0.388 &0.401 &0.367 &0.385 &0.374 &0.387 &0.387 &0.392
\\
& 336 & \underline{0.393} & \textbf{0.400} & 0.394 & 0.401 &\textbf{0.382} & \underline{0.401}
&0.390 &0.404 &0.421 &0.426 &0.399 &0.410 &0.410 &0.411 &0.424 &0.415
\\
& 720 &0.456 &\textbf{0.435} &\underline{0.457} &0.435 &\textbf{0.445} &\underline{0.435} 
&0.498 &0.482 &0.486 &0.465 &\underline{0.454} &0.439 &0.478 &0.450 &0.487 &0.450
\\

\midrule
\multirow{4}{*}{\rotatebox{90}{ETTm2}}
& 96  & \textbf{0.173} &\underline{0.254} &\underline{0.173} &\textbf{0.253} &0.174 &0.255
&0.175 &0.258 &0.189 &0.277 &0.175 &0.259 &0.187 &0.267 &0.182 &0.265
\\
& 192 & \textbf{0.237} &\textbf{0.295} &\underline{0.237} &\underline{0.295} &0.239 &0.299 
&0.237 &0.299 &0.258 &0.326 &0.241 &0.302 &0.249 &0.309 &0.246 &0.304
\\
& 336 & \textbf{0.297} &\textbf{0.336} &\underline{0.299} &\underline{0.336} &0.301 &0.340
&0.298 &0.340 &0.343 &0.390 &0.305 &0.343 &0.321 &0.351 &0.307 &0.342
\\
& 720 & 0.394 &\underline{0.393} &\textbf{0.391} &\textbf{0.392} &0.395 &0.396 
&\underline{0.391} &0.396 &0.495 &0.480 &0.402 &0.400 &0.408 &0.403 &0.407 &0.398
\\

\midrule
\multirow{4}{*}{\rotatebox{90}{ETTh1}}
& 96  & \textbf{0.368} &\underline{0.395} &0.374 &\textbf{0.387} &\underline{0.368} &0.398 
&0.375 &0.400 &0.395 &0.425 &0.414 &0.419 &0.384 &0.402 &0.386 &0.395
\\
& 192 & \underline{0.416} &\textbf{0.416} &0.429 &\underline{0.416} &\textbf{0.414} &0.420 
&0.429 &0.421 &0.448 &0.440 &0.460 &0.445 &0.436 &0.429 &0.437 &0.424
\\
& 336 & \textbf{0.431} &\textbf{0.426} &\underline{0.455} &0.430 &0.445 &\underline{0.434} 
&0.484 &0.458 &0.343 &0.390 &0.501 &0.466 &0.491 &0.469 &0.479 &0.446
\\
& 720 & \underline{0.464} &\textbf{0.459} &0.481 &0.473 &\textbf{0.444} &\underline{0.459} 
&0.498 &0.482 &0.558 &0.532 &0.500 &0.488 &0.521 &0.500 &0.481 &0.470
\\

\midrule
\multirow{4}{*}{\rotatebox{90}{ETTh2}}
& 96  &\textbf{0.277} &\textbf{0.331} &0.278 &\underline{0.332} &\underline{0.290} &0.340
&0.289 &0.341 &0.309 &0.364 &0.302 &0.348 &0.340 &0.374 &0.318 &0.363
\\
& 192 &\textbf{0.350} &\textbf{0.379} &\underline{0.350} &\underline{0.379} &0.375 &0.392
&0.372 &0.392 &0.395 &0.425 &0.388 &0.400 &0.402 &0.414 &0.401 &0.412
\\
& 336 &\textbf{0.363} &\textbf{0.398} &\underline{0.371} &\underline{0.402} &0.423 &0.435
&0.386 &0.414 &0.462 &0.467 &0.426 &0.433 &0.452 &0.452 &0.436 &0.442
\\
& 720 &\textbf{0.412} &\textbf{0.433} &0.421 &0.439 &0.443 &0.449
&\underline{0.412} &\underline{0.434} &0.721 &0.604 &0.431 &0.446 &0.462 &0.468 &0.442 &0.454
\\

\midrule
\multirow{4}{*}{\rotatebox{90}{ECL}}
& 96  &\textbf{0.146} &0.248 &\underline{0.150} &\textbf{0.242} &0.174 &0.266 
&\underline{0.153} &\underline{0.247} &0.309 &0.364 &0.181 &0.270 &0.168 &0.272 &0.201 &0.281
\\
& 192 &\textbf{0.158} &\underline{0.256} &\underline{0.163} &\textbf{0.253} &0.182 &0.273 
&0.166 &0.256 &0.175 & 0.262 &0.188 &0.274 &0.184 &0.289 &0.201 &0.283
\\
& 336 &\textbf{0.178} &\underline{0.273} &\underline{0.179} &\textbf{0.271} &0.197 &0.286 
&0.185 &0.277 &0.185 &0.278 &0.204 &0.293 &0.198 &0.300 &0.215 &0.298
\\
& 720 &\textbf{0.214} &\textbf{0.306} &\underline{0.222} &\underline{0.307} &0.236 &0.320
&0.225 &0.310 &0.220 &0.315 &0.246 &0.324 &0.220 &0.320 &0.257 &0.331
\\

\midrule
\multirow{4}{*}{\rotatebox{90}{Traffic}}
& 96  &\textbf{0.438} &0.293 &0.466 &\underline{0.286} &0.612 &0.391 
&0.462 &\textbf{0.285} &0.593 &0.378 &0.475 &0.290 &0.593 &0.321 &0.649 &0.389
\\
& 192 &\textbf{0.452} &0.299 &0.492 &0.297 &0.580 &0.368 
&0.473 &\textbf{0.296} &0.595 &0.377 &\underline{0.466} &\underline{0.296} &0.617 &0.336 &0.601 &0.366
\\
& 336 &\textbf{0.464} &0.310 &\underline{0.493} &\underline{0.298} &0.593 &0.368
&0.498 &\textbf{0.296} &0.609 &0.385 &0.482 &0.304 &0.629 &0.336 &0.609 &0.369
\\
& 720 &0.519 &0.347 &0.527 &\underline{0.318} &0.630 &0.388
&\textbf{0.506} &\textbf{0.312} &0.673 &0.418 &\underline{0.514} &0.322 &0.640 &0.350 &0.647 &0.387
\\

\midrule
\multirow{4}{*}{\rotatebox{90}{Weather}}
& 96  &\textbf{0.154} &\textbf{0.199} &0.164 &\underline{0.205} &\underline{0.162} &0.208 
&0.163 &0.209 &0.174 &0.208 &0.177 &0.218 &0.172 &0.220 &0.192 &0.232
\\
& 192 &\textbf{0.206} &\underline{0.248} &\underline{0.208} &\textbf{0.247} &0.207 &0.249 
&0.208 &0.250 &0.219 &0.250 &0.225 &0.259 &0.219 &0.261 &0.240 &0.271
\\
& 336 &0.265 &\textbf{0.289} &0.267 &0.291 &\textbf{0.263} &\underline{0.290} 
&\underline{0.263} &0.293 &0.273 &0.290 &0.278 &0.297 &0.280 &0.306 &0.292 &0.307
\\
& 720 &0.340 &0.340 &0.341 &\underline{0.339} &\underline{0.338} &0.341 
&0.344 &0.348 &\textbf{0.334} &\textbf{0.332} &0.354 &0.348 &0.365 &0.359 &0.364 &0.353
\\

\midrule
\multirow{4}{*}{\rotatebox{90}{Exchange}}
& 96  &\textbf{0.081} &\textbf{0.201} &0.088 &0.206 &\underline{0.087} &\underline{0.206} 
&0.085 &0.204 &0.091 &0.217 &0.088 &0.205 &0.107 &0.234 &0.093 &0.217
\\
& 192 &\textbf{0.176} &\textbf{0.299} &0.185 &0.304 &0.181 &\underline{0.299} 
&0.180 &0.302 &0.175 &0.310 &\underline{0.176} &0.299 &0.226 &0.344 &0.184 &0.307
\\
& 336 &0.335 &\underline{0.417} &0.336 &0.418 &0.347 &0.426
&0.361 &0.438 &\underline{0.334} &0.434 &\textbf{0.301} &\textbf{0.397} &0.367 &0.448 &0.351 &0.432
\\
& 720 &\underline{0.861} &\underline{0.698}  &0.876 &0.705 &0.995 &0.748 
&1.011 &0.744  &\textbf{0.716} &\textbf{0.674} &0.901 &0.714 &0.964 &0.746 &0.886 &0.714
\\

\bottomrule
\end{tabular}
}

\caption{Results of multivariate long-term forecasting with various prediction lengths (96, 192, 336, 720). The best and second performances are bolded and underlined, respectively.}
\label{table1}
\end{table*}

Specifically, we adopt Chebyshev polynomials \(T_{n}(x)=\cos  (n\cdot \arccos (x))\) as the functional basis to construct expressive univariate functions. The learnable univariate function \(\phi _{o}(x)\), corresponding to the \(o\)-th output neuron, is defined as a linear combination of Chebyshev polynomials:
\begin{equation}
\phi _{o}(x)= \sum_{j=1}^{D}\sum_{i=0}^{n}\Theta _{o,j,i}T_{i}(\text{Tanh} (x_{j} ))\, ,
\end{equation}

\begin{equation}
\text{KAN}(x)=\left \{ \begin{matrix}\phi _{i}(x) 
 \\\cdots
 \\\phi _{D}(x)

\end{matrix} \right \}\, ,
\end{equation}
where \(n\) denotes the highest order of the Chebyshev polynomial, and \(\Theta  \in \mathbb{R} ^{D\times D\times(n+1)} \)represents the learnable parameters.  To better model the increased complexity and variability of high-frequency components, we assign higher-order Chebyshev expansions to higher-frequency bands. Specifically, for the input feature \(f_{i} \) from the \(i\)-th frequency band, we apply a KAN transformation with order \(b+i\), where \(b\) is the beginning  polynomial order. The output is defined as:
\begin{equation}
\widehat{ f}_{i} = \text{KAN} (f_{i}, order=b+i)+ f_{i}\, ,
\end{equation}
where the residual connection helps preserve original information and stabilize training.


\subsection{Training Objective}
Following the MBS module, a head module is employed to map the feature representations of each frequency component to the target prediction length. Specifically, the head module takes the fused feature vectors as input and applies a FFN to produce the predicted wavelet coefficients with the prediction dimensionality \(d_{\text{pred}}\):
\begin{equation}
\widehat{x} _{i} = \text{FFN}_{\text{head}}(\widehat{ f}_{i}) \in \mathbb{R}^{C\times d_{\text{pred}}}\, .
\end{equation}

Then, the reconstruction of the wavelet coefficient sequence back to the time domain is performed via the inverse wavelet packet transform (IWPT) as follows:
\begin{equation}
\widehat{X}_{T}=\text{IWPT} _{\psi }(\widehat{x} _{1},\widehat{x} _{2},...,\widehat{x} _{t+T})\in \mathbb{R}^{C\times T}\, .
\end{equation}

To optimize the model parameters and ensure robustness to outliers, we adopt the SmoothL1Loss as the training objective, \ie,  
\begin{equation} \nonumber
L=
\begin{cases}
  (0.5(X_{T}-\hat{X}_{T} )^{2})/T,& \text{ if }  \left |X_{T} -\hat{X}_{T} \right |< 1  \\
  (\left | X_{T}-\hat{X}_{T} \right |-0.5)/T, & \text{ otherwise }\, .
\end{cases}
\end{equation}

\section{Experiments}

\subsection{Experimental Setup}
\paragraph{Datasets \& Evaluation Metrics.} We evaluated the performance of our WaveTuner on 8 commonly used LSTF benchmark datasets: ETT~\cite{zhou2021informer}, Exchange~\cite{lai2018modeling}, Weather~\cite{wu2021autoformer}, Electricity~\cite{wu2021autoformer}, and the Traffic~\cite{wu2021autoformer} dataset. Two commonly used metrics are used for evaluation, \ie, Mean Squared Error (MSE) and Mean Absolute Error (MAE). More details on datasets and evaluation metrics are presented in the supplementary.


\paragraph{Baselines.} We selected eight widely acknowledged SOTA models as benchmarks for comparison. These include wavelet-based model:WPMixer~\cite{murad2025wpmixer}, frequency-based models: TimeKAN~\cite{huang2025timekan},  FreTS~\cite{yi2023frequency}, time-domain models: TimesNet~\cite{wu2022timesnet},  TimeMixer~\cite{wang2024timemixer}, RLinear~\cite{li2023revisiting}, PatchTST~\cite{nie2022time}.

\paragraph{Implemention Details.} Following TimeKAN~\cite{huang2025timekan} settings, the lookback window and prediction lengths are set to $L=96$ and $T=\{96,192,336,720\}$ for all experiments. 
All models are implemented in PyTorch 2.1.2.  Except that the traffic dataset is evaluated on an H20-NVLink GPU, others are on a NVIDIA RTX 4090. More details are provided in the supplementary.


\begin{table}[t]
\centering
\setlength{\tabcolsep}{1mm}{
\begin{tabular}{c|cc|cc|cc}
\toprule
\textbf{Datasets}
& \multicolumn{2}{c|}{ETTm2} 
& \multicolumn{2}{c|}{ECL} 
& \multicolumn{2}{c}{Traffic} 
\\
\cmidrule(r){1-1} \cmidrule(r){2-3} \cmidrule(r){4-5} \cmidrule(r){6-7}
\textbf{Metrics} 
& \textbf{MSE} & \textbf{MAE} 
& \textbf{MSE} & \textbf{MAE} 
& \textbf{MSE} & \textbf{MAE} 
\\

\midrule
DWT &0.277 &0.320 &0.183 &0.277 &0.477 &0.314
\\

w/o Ada &0.282 &0.323 &0.180 &0.275 &0.506 &0.321
\\

w/o WE &0.277 &0.320 &0.195 &0.280 &0.514 &0.328
\\

MLPs &0.278 &0.321 &0.182 &0.277 &0.495 &0.321
\\

FLOK &0.279 &0.323 &0.180 &0.274 &0.485 &0.317
\\

FHOK &0.280 &0.324 &0.177 &0.272 &0.487 &0.315
\\

\textbf{WaveTuner} &\textbf{0.275} &\textbf{0.319} &\textbf{0.174} &\textbf{0.271} &\textbf{0.469} &\textbf{0.311}
\\

\bottomrule

\end{tabular}
}

\caption{Ablation of WaveTuner, averaged over $T 
\in \{96, 192, 336, 720\}$.}
\label{table2}
\end{table}

\begin{figure}[!t]
\centering
    \begin{subfigure}{0.49\linewidth}
        \centering
        \includegraphics[width=1\linewidth]{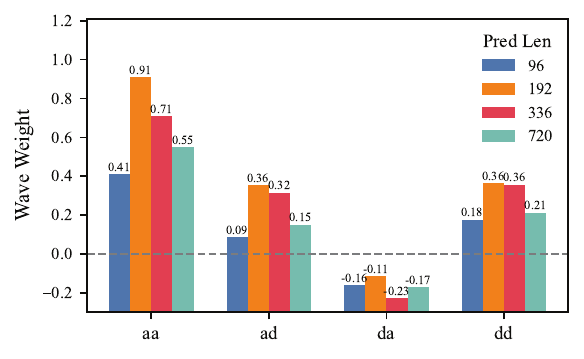}
        \caption{Learned frequency weights of different horizons}
        \label{ETTm1_weigth_dlen}
    \end{subfigure}
    \centering
    \begin{subfigure}{0.49\linewidth}
        \centering
        \includegraphics[width=1\linewidth]{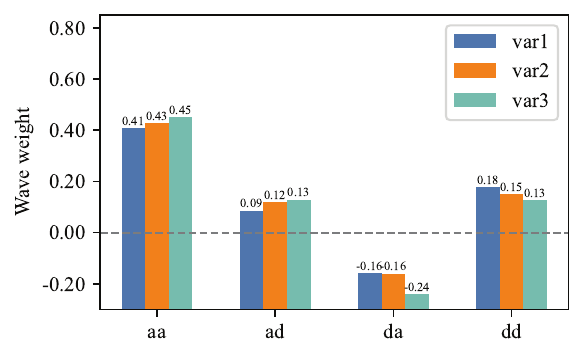}
        \caption{Learned weights of different variables}
        \label{ETTm1_weigth_96}
    \end{subfigure}
    \centering

\caption{
Visualization of learned weight.
}
\label{fig5}
\end{figure}

\begin{figure*}[!t]
\centering
\begin{subfigure}{0.33\textwidth}
    \centering
    \includegraphics[width=1\textwidth]{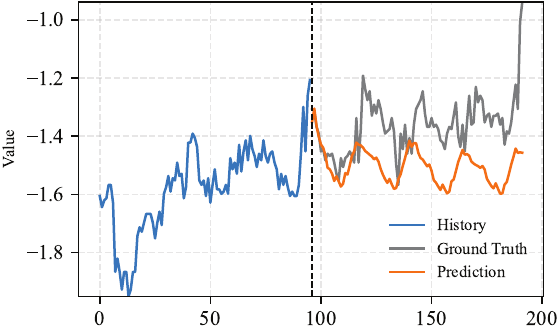}
    \subcaption{WPMixer}
\end{subfigure}
\begin{subfigure}{0.33\textwidth}
    \centering
    \includegraphics[width=1\textwidth]{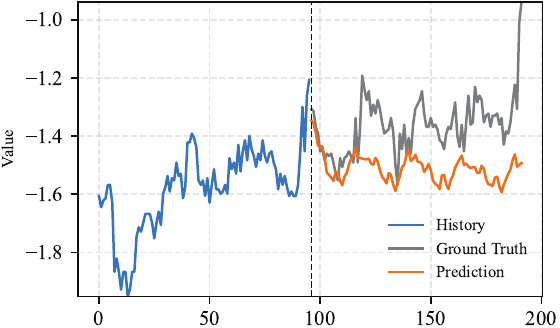}
    \subcaption{TimeKAN}
\end{subfigure}
\begin{subfigure}{0.33\textwidth}
    \centering
    \includegraphics[width=1\textwidth]{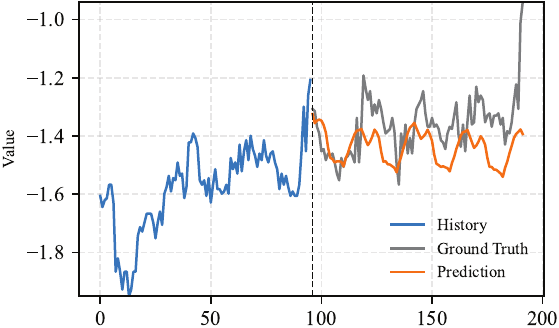}
    \subcaption{WaveTuner}
\end{subfigure}
\caption{Visualization of predictions on the ETTh1 dataset with lookback and horizon length as 96.}
\label{fig3}
\vspace{-2mm}
\end{figure*}

\subsection{Main Results}
Table \ref{table1} presents the forecasting results on all datasets. WaveTuner consistently outperforms existing baselines, validating the effectiveness of integrating fine-grained frequency decomposition and frequency-aware specialization for long-term time series forecasting. In particular, it achieves notable improvements over channel-independent models such as WPMixer, TimeKAN and PatchTST, suggesting that explicitly modeling inter-variable dependencies in multi-resolution contributes significantly to performance gains. WaveTuner consistently outperforms existing baselines, validating the effectiveness of integrating fine-grained frequency decomposition and frequency-aware specialization for long-term time series forecasting. Although both WaveTuner and TimeKAN adopt KAN-based modeling, the superior performance of WaveTuner highlights the benefit of coupling KAN with wavelet-based subband decomposition and specialization. Compared to frequency-domain models like FreTS, WaveTuner's multi-resolution formulation offers richer frequency representations, enabling better capture of complex periodic structures. Moreover, methods like RLinear that rely on static linear mappings without decomposition fall short in modeling hierarchical temporal dependencies, further demonstrating the advantages of frequency-aware architectures.

\begin{figure}[t]
\centering
\includegraphics[width=0.95\columnwidth]{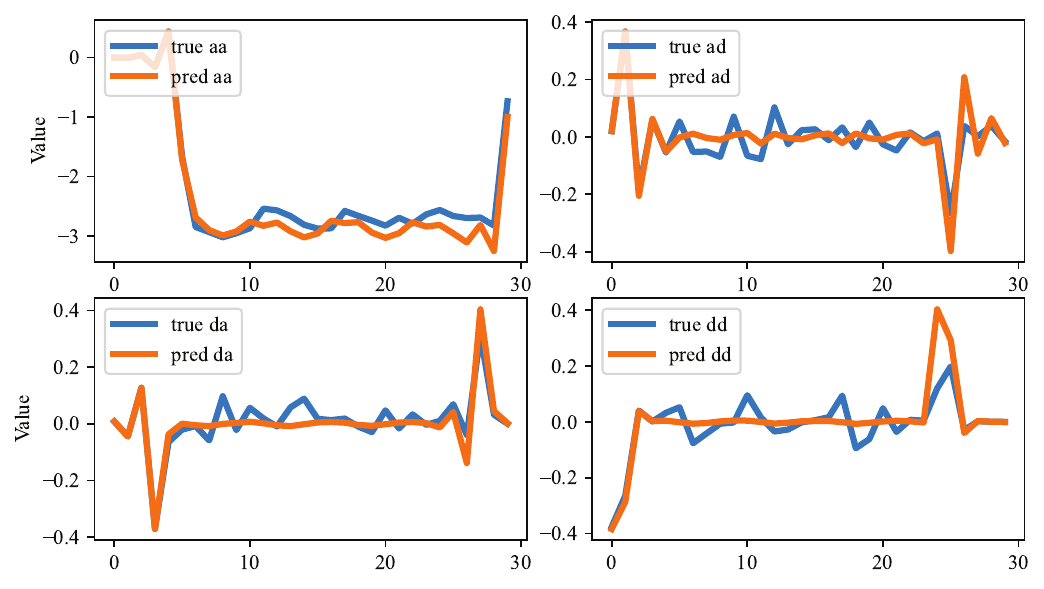} 
\caption{Decomposition visualization of the prediction. 
}
\label{fig4}
\vspace{-2mm}
\end{figure}

\subsection{Model Analysis}


\paragraph{Tuning Capability of Adaptive Wavelet Refinement.}

We perform ablation studies to validate the components within the Adaptive Wavelet Refinement module, including Wavelet Packet Decomposition (WPD), adaptive weighting, and the Wave Embedding (WE) mechanism. The results are presented in Table~\ref{table2}. Replacing WPD with standard Discrete Wavelet Transform (DWT) leads to degraded performance, highlighting the benefit of fine-grained frequency decomposition. Removing the adaptive weighting module (w/o Ada) also downgrades accuracy, confirming its role in dynamically emphasizing relevant frequency bands. Furthermore, substituting the WE module with a simple MLP (w/o WE) results in performance drops, demonstrating the effectiveness of modeling inter-variable dependencies across multiple resolutions.

\paragraph{Single- vs. Multi-Branch Specialization.} Three variants are designed to demonstrate the effective of this module: (1) \textbf{MLPs}: replace each KAN with an MLP; (2) \textbf{FLOK}: use \textbf{F}ixed \textbf{L}ow-\textbf{O}rder (order 2) \textbf{K}ANs for all subbands; (3) \textbf{FHOK}: use \textbf{F}ixed \textbf{H}igh-\textbf{O}rder (order 5) \textbf{K}ANs. As shown in Table~\ref{table2}, our specialization module achieves the best performance. It outperforms MLPs by leveraging order adaptivity, and exceeds both \textbf{FLOK} and \textbf{FHOK}, demonstrating the benefit of assigning appropriate functional complexity to different frequency bands.

\paragraph{Frequency Band Weight Distributions.} 
Figure~\ref{fig5} (a) illustrates the learned frequency weights for a single variable under prediction lengths of 96, 192, 336, and 720 on ETTm1. With a decomposition depth of 2, the input is divided into four subbands (\(aa, ad, da, dd\)). Despite identical inputs, the model assigns different importance to each subband across forecasting horizons, indicating step-dependent frequency selection. Figure~\ref{fig5} (b) shows the frequency weights for three representative variables under a prediction length of 96. The distinct weighting patterns across variables indicate variable-specific frequency preferences. These results demonstrate that WaveTuner’s adaptive wavelet refinement effectively captures task-relevant frequency features, improving model flexibility and forecasting accuracy. 

\paragraph{Visualization of Prediction.} Figure~\ref{fig3} presents predictions for a sample from the ETTh1 dataset. WaveTuner is compared with two representative baselines: WPMixer (wavelet-based) and TimeKAN (frequency-based). As shown in the figure, WaveTuner generates more accurate and smoother forecasts. Specifically, it better preserves informative frequency components while mitigating overfitting to noise, outperforming both baselines. To further interpret the prediction behavior, we visualize in Figure~\ref{fig4} the wavelet coefficients corresponding to WaveTuner’s output in Figure~\ref{fig3}. The four plots present the predicted versus ground truth coefficients across four subbands (\(aa\), \(ad\), \(da\), \(dd\)) from a level-2 wavelet decomposition. We observe that low-frequency bands (\eg, \(aa\)) capture major trends and are accurately estimated, while high-frequency bands (\eg, \(dd\)) exhibit flatter structures, and the model avoids fitting spurious fluctuations. Despite using high-order KANs for high-frequency modeling, the network adaptively suppresses unnecessary complexity when signal variance is low. This selective expressiveness demonstrates WaveTuner’s frequency-aware design—preserving detail where informative, and promoting generalization where appropriate—aligning well with our core motivation.

\begin{figure}[!t]
\centering
\includegraphics[width=0.8\columnwidth]{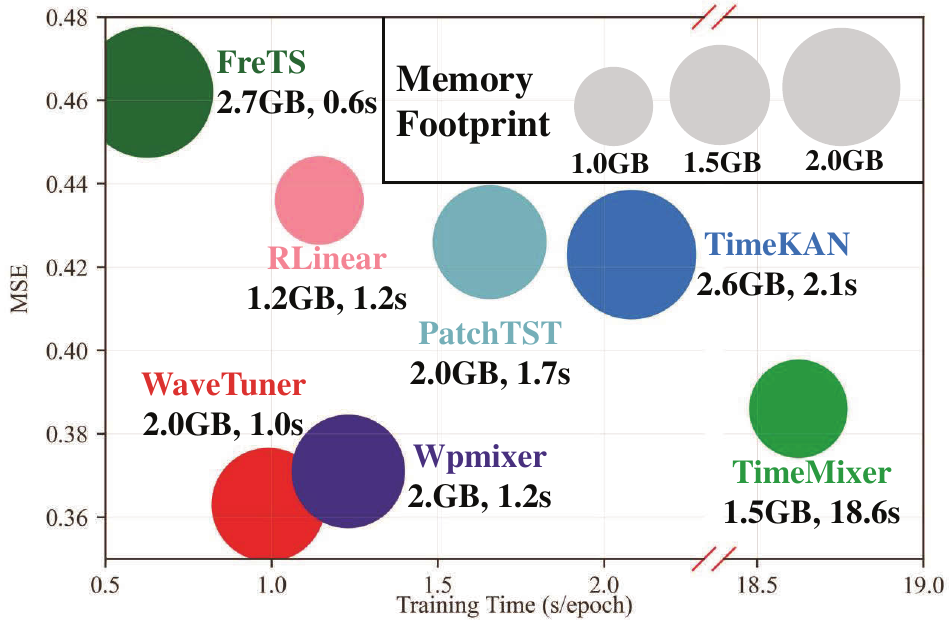} 
\caption{Effectiveness and efficiency comparison.}
\label{fig6}
\vspace{-2mm}
\end{figure}


\paragraph{Efficiency Analysis.} The overall computational complexity of WaveTuner is determined by its two core components: Adaptive Wavelet Refinement and Multi-Branch Specialization. Specifically, Wavelet Packet Transform incurs a cost of \(\mathcal{O}(C\cdot L\cdot m )\), Adaptive Weighting \(\mathcal{O}(C\cdot 2^{m} )\), Wave Embedding \(\mathcal{O}(L\cdot C \cdot d)\), where \(C\), \(L\), \(m\), and \(d\) denote the number of variables, sequence length, decomposition levels, and embedding dimension, respectively. Multi-Branch Specialization introduces a cost of \(\mathcal{O}(L\cdot d^{2} \cdot K)\), with \(K\) are constants in practice, the total complexity scales linearly with respect to both the number of variables and the sequence length: \(\mathcal{O}(C\cdot L)\). Figure~\ref{fig6} shows the empirical comparison on ETTh2 with a history window of 96 and a prediction length of 336. WaveTuner not only delivers the best forecasting accuracy but also surpasses WPMixer in terms of computational efficiency.

\section{Conclusion}
In this paper, we propose WaveTuner, a novel frequency-aware forecasting framework that integrates adaptive wavelet refinement with multi-branch specialization to capture temporal dynamics across multiple frequency bands. The refinement module conducts fine-grained wavelet packet decomposition and adaptively weights subbands based on task- and variable-specific properties. The specialization module assigns tailored learners to each subband, enabling frequency-aware and disentangled representations. This design allows the model to flexibly capture both global trends and local variations across diverse temporal patterns. Extensive experiments show that WaveTuner outperforms strong baselines, validating the effectiveness of combining adaptive frequency modeling with specialization.

\bibliography{aaai2026}

\end{document}